\documentclass[10pt,twocolumn,letterpaper]{article}

\usepackage[pagenumbers]{iccv} 

\usepackage{amsthm}
\newtheorem{assumption}{Assumption}
\newtheorem{theorem}{Theorem}
\theoremstyle{remark}
\newtheorem{remark}{Remark} 

\definecolor{iccvblue}{rgb}{0.21,0.49,0.74}
\usepackage[pagebackref,breaklinks,colorlinks,allcolors=iccvblue]{hyperref}

\usepackage{multirow}
\usepackage{amssymb}
\usepackage{pifont}
\usepackage{soul}
\usepackage{algpseudocode}
\usepackage{algorithm}

\newcommand{\res}[1]{$#1$}
\newcommand{\resg}[1]{$\color{gray}#1$}
\newcommand{\resb}[1]{$\mathbf{#1}$}
\newcommand{\cmark}{\ding{51}}%
\newcommand{\xmark}{\ding{55}}%
\newcommand{\ic}[1]{\textcircled{\raisebox{-0.9pt}{#1}}}

\def\paperID{6502} 
\def\confName{ICCV}
\def\confYear{2025}

\title{Closer to Reality: Practical Semi-Supervised Federated Learning for\\ Foundation Model Adaptation}

\author{Guangyu Sun\textsuperscript{1,2}\thanks{Equal Contribution. Work done during an internship at Sony AI.}, Jingtao Li\textsuperscript{1*}, Weiming Zhuang\textsuperscript{1}, Chen Chen\textsuperscript{1}, Chen Chen\textsuperscript{2}, Lingjuan Lyu\textsuperscript{1}\\
\\
\textsuperscript{1} Sony AI\\
\textsuperscript{2} University of Central Florida, USA
}

\begin{document}
\maketitle
\begin{abstract}
Foundation models (FMs) exhibit remarkable generalization but require adaptation to downstream tasks, particularly in privacy-sensitive applications. Due to data privacy regulations, cloud-based FMs cannot directly access private edge data, limiting their adaptation. Federated learning (FL) provides a privacy-aware alternative, but existing FL approaches overlook the constraints imposed by edge devices—namely, limited computational resources and the scarcity of labeled data.
To address these challenges, we introduce Practical Semi-Supervised Federated Learning (PSSFL), where edge devices hold only unlabeled, low-resolution data, while the server has limited labeled, high-resolution data. In this setting, we propose the Federated Mixture of Experts (FedMox), a novel framework that enhances FM adaptation in FL. FedMox tackles computational and resolution mismatch challenges via a sparse Mixture-of-Experts architecture, employing a spatial router to align features across resolutions and a Soft-Mixture strategy to stabilize semi-supervised learning.
We take object detection as a case study, experiments on real-world autonomous driving datasets demonstrate that FedMox effectively adapts FMs under PSSFL, significantly improving performance with constrained memory costs on edge devices. Our work paves the way for scalable and privacy-preserving FM adaptation in federated scenarios.
\end{abstract}    
\section{Introduction}
\label{sec:intro}


Foundation models (FMs) have emerged as powerful tools in modern machine learning, demonstrating remarkable generalization across diverse tasks~\cite{bommasani2021opportunities, firoozi2023foundation, huang2023applications, kawaharazuka2024real}. This shift has driven research into \textit{an era of foundation models}, where large-scale, pre-trained models serve as versatile backbones for a wide range of downstream applications. However, these models are typically trained on publicly available large-scale datasets~\cite{oquab2023dinov2, radford2021learning, caron2021emerging, chen2023vision}, while sensitive or private data is often inaccessible or excluded, leading to performance degradation when applied to domain-specific tasks involving such data. Consequently, further adaptation of FMs is essential to optimize their effectiveness for real-world applications.

\begin{figure}[t] 
\centering 
\includegraphics[width=\linewidth]{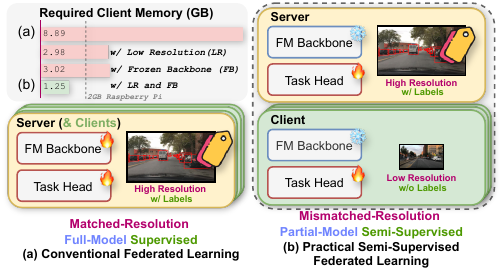} 
\caption{(a) Conventional federated learning focuses on an ideal scenario where the clients have labeled data and sufficient resources to train the full model with the same resolution as the server. (b) We study a more {practical and challenging} scenario, where clients only have unlabeled data and are only able to train the partial model with a frozen backbone (FB) and a lower resolution (LR), significantly reducing the required client memory to enable training on edge devices, \eg, a 2GB Raspberry Pi.} 
\label{fig:setting} 
\vspace{-3mm}
\end{figure}


However, the adaptation of FMs in privacy-sensitive settings is increasingly constrained by growing concerns around data privacy and stringent regulations such as the General Data Protection Regulation (GDPR)~\cite{GDPR2016} and the California Consumer Privacy Act (CCPA)~\cite{CCPA2018}. Collecting and aggregating sensitive data to fine-tune FMs is becoming more challenging, necessitating privacy-preserving training paradigms. {Federated learning} (FL)~\cite{fedavg} offers a promising approach by enabling collaborative model training across decentralized data sources while ensuring that raw data remains localized on clients. By allowing multiple clients (\eg, edge devices) to compute local model updates 
for aggregation on the central server, FL provides a privacy-preserving mechanism to adapt FMs while mitigating data governance concerns~\cite{hallaji2024decentralized}.

Despite its promise, deploying FMs in FL presents significant challenges, primarily due to \textit{the scarcity of labeled data} and \textit{the computational constraints} of edge devices. Conventional FL research has predominantly focused on smaller models trained in a fully supervised setting, assuming that clients possess sufficient resources for full-model training and access to labeled data~\cite{fedavg, sun2022conquering, fedprox}, as illustrated in Figure~\ref{fig:setting}(a). In contrast, real-world scenarios often lack abundant labeled data, with only the central server possessing a limited labeled dataset while edge devices hold unlabeled data~\cite{yu2020bdd100k, han2021soda10m}. While recent efforts have begun exploring \textit{semi-supervised} or \textit{unsupervised FL}, these approaches have been largely restricted to \textit{simple classification tasks with small models}~\cite{diao2022semifl, lin2021semifed}. More complex tasks requiring structured outputs, such as \textit{object detection}, remain under-explored~\cite{kim2024navigating}. These gaps underscore the necessity for {semi-supervised federated learning (SSFL) methods tailored for more complex tasks and large models}.

Beyond the challenge of labeled data scarcity, computational constraints present a critical bottleneck. Advances in model compression have made it feasible to \textit{deploy} FMs for efficient inference on edge devices~\cite{hershcovitch2024lossless, ye2024once}. However, \textit{training} an FM requires storing full-precision intermediate features for backpropagation, making model compression techniques ineffective, and exceeding the memory capacity of most edge devices. 
Additionally, {image resolution significantly impacts memory consumption}. For instance, object detection models require high-resolution inputs to detect small objects, leading to substantially larger feature maps and intensifying memory demands. Unfortunately, this factor has been largely overlooked in prior SSFL studies~\cite{diao2022semifl, lin2021semifed}.


To address these constraints,  adaptations are required. Since the majority of memory is consumed by {intermediate feature storage in the FM backbone}, a viable solution is to freeze the backbone and fine-tune only the \textit{task head}, \ie, the module responsible for processing high-level representations into final predictions. Additionally, reducing the input resolution on edge devices can further decrease feature map size, making training feasible under resource constraints.

These necessary compromises motivate us to explore a practical setting for FM adaptation in FL with constraints on data label availability and computation.
We introduce this setting as \textbf{Practical Semi-Supervised Federated Learning (PSSFL)}, characterized by \textit{semi-supervised learning} with a \textit{frozen backbone} under \textit{mismatched resolutions}, \ie, high-resolution images on the server and low-resolution images on clients. In this paper, we take more complex \textit{object detection} as 
an example. 
As illustrated in Figure~\ref{fig:setting}, training a Faster R-CNN~\cite{ren2016faster} model with a batch size of 2 in high resolution ($1280\times720$) under conventional FL requires \textbf{8.89GB} of memory, exceeding the capacity of many edge devices. In contrast, training with low resolution ($640\times360$) and a frozen backbone reduces client-side memory consumption to just \textbf{1.25GB}, making it a practical solution for real-world applications. For example, a commonly used edge device, a \textit{2GB Raspberry Pi}, can feasibly accommodate this reduced computational demand.

However, adapting to these constraints introduces additional challenges:
\textit{(a)~Limited Learning Capacity.} 
With a frozen backbone, the task head alone may lack sufficient generalization ability, leading to performance degradation across diverse client data distributions.
\textit{(b)~Resolution Mismatch.} 
The discrepancy between high-resolution server training and low-resolution client training introduces a learning gap, requiring mechanisms to effectively align feature representations across different resolutions.
\textit{(c)~Sequential Updates.} 
Unlike centralized semi-supervised learning, where supervised and unsupervised updates occur in parallel, in FL, they are conducted sequentially, \ie, supervised learning on the server and unsupervised learning on clients, leading to instability in the training process.

In response to these challenges, we propose a novel framework based on a Mixture of Experts (MoE) architecture, FedMox (\textbf{Fed}erated \textbf{M}ixture \textbf{o}f E\textbf{x}perts). The MoE structure enhances the task head’s capacity, compensating for the frozen backbone. To maintain efficiency, we adopt \textit{sparse expert activation}, allowing only a subset of experts to be trained per client. To bridge resolution mismatches, we introduce a \textit{spatial router} that directs different feature regions to specialized experts. Furthermore, to stabilize knowledge aggregation under sequential updates, we develop a \textit{Soft-Mixture} strategy that effectively balances supervised training on the server with unsupervised updates on clients. Our {contributions} are summarized as follows:
\begin{itemize}
    \item To the best of our knowledge, this paper is the \textit{first} work in FL to study \textit{FM adaption} with \textit{practical constraints} on a \textit{complex task} (\ie, object detection).
    \item We propose a novel and practical semi-supervised FL setting, PSSFL, which integrates FM-based frozen backbones and mismatched resolution inputs, addressing both computational and labeling constraints on edge devices.
    \item We propose a novel framework, FedMox, to overcome key challenges in PSSFL, including limited learning capacity, resolution mismatch, and sequential updates. Experimental evaluations on real-world autonomous driving datasets demonstrate the effectiveness of our approach, paving the way for future FL research in the era of foundation models.
\end{itemize}

\section{Related Work}

\noindent \textbf{Federated Learning} (FL) has emerged as a promising privacy-preserving training paradigm~\cite{fedavg, kairouz2021advances,wang2021field}. However, previous work has primarily focused on classification tasks, with most experiments limited to small models (\eg, 2-layer MLPs) and datasets (\eg, MNIST~\cite{lecun1998gradient}, CIFAR-10/100~\cite{krizhevsky2009learning}), as the primary focus has been on optimization in FL. Recently, studies have started exploring the potential of incorporating FMs into FL~\cite{zhuang2023foundation, sun2022conquering, sun2023fedperfix, sun2024towards}, though most are still confined to supervised learning settings. This paper investigates a practical setting for FM adaptation in FL, addressing the constraints of labels and computation on clients. 
Unlike prior work, we extend our study to the more complex task of object detection, providing insights into adapting FMs to resource-limited, semi-supervised and resolution-mismatch federated learning scenarios.


\noindent \textbf{Semi-supervised Federated Learning.}
Semi-supervised learning (SSL) is an important and practical training paradigm, particularly for tasks that require dense annotations, such as object detection~\cite{yang2022survey,wang2023semi,shehzadi2024semi,xu2021end}. 
In FL, existing semi-supervised approaches mostly focus on classification tasks. One line of work, such as SemiFed~\cite{lin2021semifed}, investigates scenarios where each client has a different proportion of labeled and unlabeled data, exploring how such data distributions affect model performance in a FL setting. Another relevant study, SemiFL~\cite{diao2022semifl}, examines a more practical FL scenario where clients hold only unlabeled data, and the server holds the limited labeled data, effectively addressing common data constraints in real-world FL applications. The most closely related work to ours is FedSTO~\cite{kim2024navigating}, which investigates semi-supervised federated object detection (SSFOD) with a lightweight Yolov5 model~\cite{jocher2022ultralytics} under matched resolution. Our paper extends this line of research with more realistic constraints in the era of foundation models.


\noindent \textbf{Mixture of Experts.} Decades ago, MoE was proposed for composing separate networks~\cite{jacobs1991adaptive}. Recently, MoE shines in the design for Large Language Models (LLMs)~\cite{jiang2024mixtral}. In FL, several attempts are made for different purposes. To address the data heterogeneity challenges for classification tasks, several methods provide a theoretical analysis of MoE~\cite{reisser2021federated} or use it for personalization~\cite{dun2023fedjets, yi2024fedmoe, luo2025mixture, mei2024fedmoe}. Most existing works leverage MoE with a traditional global routing strategy since they focus on inputs with the same resolution for classification. 
In this paper, we leverage MoE to enhance the model capability when the backbone is frozen and design a spatial sparse routing strategy to address the mismatched resolution and limited computation challenges.




\begin{figure}[t] 
\centering 
\includegraphics[width=\linewidth]{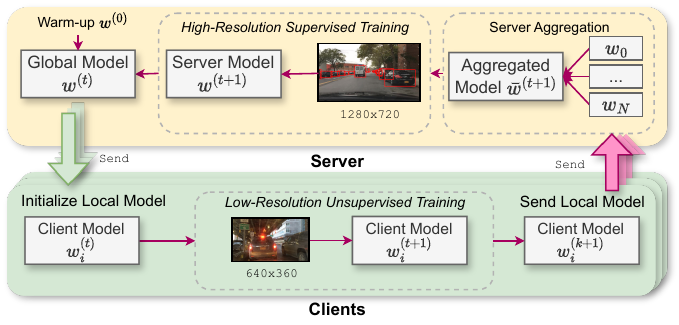} 
\vspace{-5mm}
\caption{\textbf{Overview of our proposed Practical Semi-supervised Federated Learning setting.} After training with the high-resolution labeled data on the server as a \textit{warm-up}, the global model $w^{(t=0)}$ will be sent to clients and initialize client models $w_i^{(t)}=w^{(t)}$ for \textit{low-resolution unsupervised training}. 
After local training, the updated client models $w_i^{(t+1)}$ will be returned to the server for aggregation. The server will aggregate them into $\bar w^{(t+1)}$, and it will further be trained with the \textit{high-resolution labeled data} on the server as the global model for the next round.}
\label{fig:overall} 
\vspace{-5mm}
\end{figure}

\section{Method}

\subsection{The Problem: PSSFL} \label{sec:problem}

In our PSSFL setting, we assume that there is one server and $N$ clients. Given an FM, we will freeze its backbone, \ie, the encoder that encodes input images to high-level features, 
and train a task head $w$ (\ie, Faster-RCNN head~\cite{ren2016faster} for \textit{object detection}) as the \textit{global model} in the FM adaptation. Note that the backbone parameters will only be sent once to each client before the FL starts.

The server holds $n_s$ labeled samples as $\mathcal{D}^s$, while each client $i$ holds $\mathcal{D}_i^u$ unlabeled samples.
The total number of samples $n=\sum_{i=1}^{N}n_i$, where $n_i = |\mathcal{D}_i^u|.$ 
To guarantee the privacy-preserving principle of FL, we ensure \textit{there is no overlapping between the server and clients, and between different clients.}
Before the FL process, the server performs a warm-up training with $\mathcal{D}^s$ in a high-resolution training setting for $T_w$ epochs and gets the initial global model $w^{(0)}$. 

During each round $t$ of the FL process, the server will randomly select $M = r\cdot N$ online clients per round with a ratio $r$ and send the global model to the selected clients. After receiving the global model, the client will conduct unsupervised training with its own data in a low-resolution training setting to reduce memory costs.

After local training, the client model will be sent back to the server for aggregation. The server will aggregate the client models into the global model with an aggregation algorithm (\eg, FedAvg~\cite{fedavg}).
Then the server will conduct server-side training with high-resolution images on the aggregated model $\bar w^{(t+1)}$ with $\mathcal{D}^s$. The final global model is noted as $w^{(t+1)}$.
This process, as shown in Figure~\ref{fig:overall}, will be repeated for $T$ rounds. A detailed algorithm of PSSFL is provided in the \textit{{supplementary}}.

\noindent\textbf{Evaluation.} The global model will be evaluated with a hold-out test set in a high-resolution setting where the data are from both the server and client data distributions.

\subsection{The Framework: FedMox}

Our proposed PSSFL setting presents several key challenges: \textit{limited learning capability}, \textit{mismatched resolution}, and \textit{sequential updates}.

\begin{figure}[t] 
\centering 
\includegraphics[width=\linewidth]{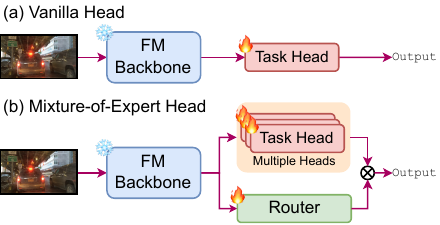} 
\vspace{-5mm}
\caption{(a) The learning capability of the FM is compromised due to the frozen backbone. A single task head lacks the flexibility to handle diverse data. (b) To compensate for the compromised flexibility, we use a Mixture-of-Expert Structure with multiple heads to boost the learning capability.} 
\label{fig:head} 
\vspace{-5mm}
\end{figure}

Limited learning capability arises due to the frozen backbone, which restricts parameter updates and demands a more powerful task head to effectively learn from diverse data.
Mismatched resolution refers to the difference in image resolutions between the server and clients due to memory constraints on edge devices, requiring the model to adapt to both small and large input sizes during training. Lastly, SSFL presents a unique challenge: sequential updates. Unlike a centralized setting, where supervised and unsupervised learning occur \textit{in parallel} in each training step, SSFL separates these processes: unsupervised learning takes place on the clients, followed by supervised learning on the server, resulting in an imbalance between supervised and unsupervised learning.

To tackle these challenges, we introduce a novel framework, FedMox, which integrates solutions at levels of both the model architecture and aggregation. In Sec.~\ref{sec:moe}, we present a \textit{Mixture-of-Experts (MoE) approach} to enhance task head learning capability and handle mismatched resolution. In Sec.~\ref{sec:sharing}, we propose a \textit{Soft-Mixture strategy} that mitigates imbalances caused by sequential updates, enabling more effective knowledge transfer in SSFL.

\subsection{Spatial Sparse Mixture of Experts}\label{sec:moe}

When training a task head with a frozen backbone, 
it lacks the flexibility required to handle diverse, unlabeled client data.
To enhance learning capability, we use a Mixture of Experts (MoE) structure with multiple task heads to compensate for the capacity lost due to the frozen backbone, as shown in Figure~\ref{fig:head}.

\begin{figure}[t] 
\centering 
\includegraphics[width=0.85\linewidth]{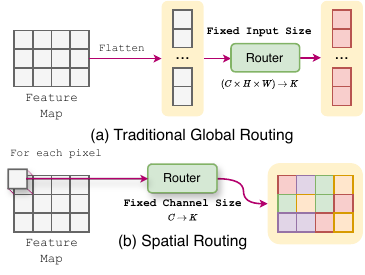} 
\vspace{-5mm}
\caption{(a) Traditional global routing will flatten the feature map and get a consistent routing map for the entire input, requiring a fixed size of height ($H$), width ($W$) and channel ($C$), which is not applicable under different resolutions. (b) Our spatial routing will route each \textit{pixel} of the feature map, requiring only a fixed size on the channel dimension ($C$), which is compatible with different resolutions. Each color indicates a routed expert $E_i \in [K]$.} 
\label{fig:routing} 
\vspace{-5mm}
\end{figure}

Traditionally, in a MoE layer with $K$ experts, the output for an input $\mathbf{x}$ is defined as:
\begin{equation}
\textstyle
    \text{MoE}(\mathbf{x}) = \sum_{i=1}^K R(\mathbf{x})_i \cdot E_i(\mathbf{x}),
\end{equation}
where $R$ is the \textit{router} (or gating network) that assigns inputs to specific experts, and $E_i(\cdot)$ denotes the $i$-th expert~\cite{jacobs1991adaptive}.

Based on the standard MoE, we aim to address two primary challenges in our proposed method: \textit{Mismatched resolution} between the server and clients and \textit{constrained computational resources} on clients. To address these needs, we design a \textit{spatial} and \textit{sparse} MoE that adapts to varying resolutions while meeting the computational constraints.

Mismatched resolutions not only impact image quality but also influence model architecture, as inputs with different resolutions produce feature maps of \textit{different sizes}. 
Although the task head may handle images of different resolutions, traditional global routers, which are usually linear layers with a fixed input dimension, typically cannot handle them~\cite{jacobs1991adaptive}.
As shown in Figure~\ref{fig:routing}, traditional global routers, designed for fixed-dimension features and routing into a fixed expert routing map for the entire feature map, struggle to handle this variation. To overcome this, we adopt a spatial MoE design inspired by MoE architectures in Natural Language Processing (NLP)~\cite{jiang2024mixtral}, where each token is routed individually. Here, each pixel in the feature map is routed independently, similar to the approach in spatial MoE models~\cite{dryden2022spatial}. For input features $\mathbf{x} \in \mathbb{R}^{C \times H \times W}$ derived from images of different resolutions, the router acts as a $1 \times 1$ convolution along the channel dimension, which only requires a fixed channel dimension of each pixel, producing a routing map $\mathbf{m} \in \mathbb{R}^{K \times H \times W}$. Such a design maintains the spatial consistency for each expert across different resolutions, which is discussed in Sec.~\ref{sec:moe_dis} and visualized in Figure~\ref{fig:expert}.

Directly training the MoE structure introduces several times of overhead in backpropagation. Hence, we employ a sparse routing strategy to keep the computational cost on par with training a single expert, which is demonstrated in Sec.~\ref{sec:compute} and shown in Figure~\ref{fig:compute}.
Specifically, we only activate the \textit{top-1} expert for each input location. Formally, for an input $\mathbf{x} \in \mathbb{R}^{C \times H \times W}$, the router is parameterized by $\mathbf{r} \in \mathbb{R}^{K \times 1 \times 1}$ and generates the routing map $\mathbf{m}$ as follows:
\begin{equation}
    \mathbf{m} = G(\mathbf{x}) = \sigma(\mathbf{r} \circledast \mathbf{x}), \mathbf{m} \in \mathbb{R}^{K \times H \times W}
\end{equation}
where $\sigma$ denotes the \textit{hard-max} operation, implementing top-1 routing, and $\circledast$ represents the convolution operation. 
This spatial sparse routing strategy allows FedMox to efficiently adapt to different resolutions while preserving computational efficiency.

\subsection{SSFL with Soft Mixture}\label{sec:sharing}
\begin{figure}[t] 
\centering 
\includegraphics[width=\linewidth]{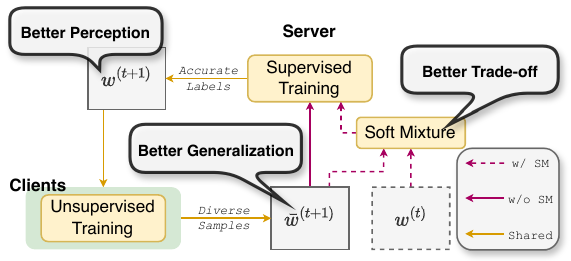} 
\caption{\textbf{Overview of the Soft Mixture (SM) strategy.} Sequential updates in SSFL cause fluctuation: the server model \(w \) is trained towards better perception, while the client-aggregated model \( \bar{w} \) is trained towards better generalizability. Soft Mixture combines the previous server model \( w^{(t)} \) with the current aggregated model \( \bar{w}^{(t+1)} \), achieving a balanced trade-off that leverages both supervised and unsupervised signals.
} 
\label{fig:soft} 
\vspace{-5mm}
\end{figure}

In a centralized semi-supervised setting, parallel updates maintain a balance between better perception when trained on labeled samples and better generalization when trained on large quantities of unlabeled samples. However, in SSFL, this balance is disrupted due to the sequential nature of updates. 
The models at various stages of training (\ie, server model $w$ after server's supervised training and aggregated model $\bar w$ after clients' unsupervised training) are trained with different objectives, as illustrated in Figure~\ref{fig:soft}. Those objectives have different generalizability error bounds~\cite{chor2023more} of expected risk $\mathcal{L}$ on test set $\mathcal{D}_t$ as
\begin{equation}
\label{eq:sup}
\textstyle
\mathcal{L}(w; \mathcal{D}_t) \le \hat{\mathcal{L}}(w) + d_{\mathcal{H}}(\mathcal{D}_s,\mathcal{D}_t) + \mathcal{O}\left( \frac{1}{\sqrt{n_s}} \right);
\end{equation}
\begin{equation}
\label{eq:unsup}
\textstyle
\mathcal{L}(\bar w; \mathcal{D}_t) \le \hat{\mathcal{L}}(\bar w) + d_{\mathcal{H}}(\mathcal{D}_u,\mathcal{D}_t) + \mathcal{O}\left( \frac{1}{\sqrt{n}} \right) + \epsilon_p,
\end{equation}
where $d_{\mathcal{H}}(\cdot,\cdot)$ is the $\mathcal{H}$-divergence, capturing the distribution shift, and $\epsilon_p$ is the pseudo-labeling error in unsupervised learning.

Considering $\mathcal{D}_u$ is more diverse and $n >> n_s$, it is safe to assume $d_{\mathcal{H}}(\mathcal{D}_u,\mathcal{D}_t) < d_{\mathcal{H}}(\mathcal{D}_s,\mathcal{D}_t)$. Hence, the server model $w$ is towards \textit{better perception} due to zero pseudo-labeling error, while the aggregated model $\bar w$ is towards \textit{better generalizability} due to a smaller $\mathcal{H}$-divergence.


That is, after unsupervised training on clients, the aggregated model tends to be trained towards greater generalizability but often at the cost of accuracy, as it lacks a strong, consistent supervised signal. Conversely, after supervised training on the server, the model shows better perception but loses some of the generalization benefits gained from the client-side updates.

To obtain a better trade-off among them within the federated setting, we propose \textit{Soft Mixture}, which derives a more balanced model that leverages both supervised and unsupervised learning signals. Specifically, we perform an additive mixing on the parameters of the previous-round server model with the current-round aggregated model to mitigate the fluctuation introduced by sequential updates.

Formally, the soft-mixed model is obtained by:
\begin{equation}
    \bar w^{(t+1)} = \alpha \cdot w^{(t)} + (1 - \alpha) \cdot \bar w^{(t+1)},
\end{equation}
where $\alpha\in [0,1]$ is a hyperparameter that controls the balance between the previous-round server model \( w^{(t)} \) and the current-round aggregated client model \( \bar w^{(t+1)} = \sum_{i=1}^{N}\frac{n_i}{n}w_i^{(t)} \). By adjusting $\alpha$, we can achieve a better trade-off between accuracy and generalizability, ensuring a more robust federated model. A more detailed proof of the existence of an optimal $\alpha^*$ is given in the \textit{{supplementary}}.



The Soft Mixture strategy enables SSFL to better handle the inherent fluctuation of sequential updates, achieving a more stable convergence and improved overall performance in semi-supervised federated scenarios.

\section{Experiments}

\begin{table*}[t]
\centering
\caption{\textbf{Evaluation on BDD100K dataset} of supervised centralized learning (SCL)
and semi-supervised learning (SSFL) methods with mAP@50. The performance on the entire test set (Total) and separate performance on each domain are reported. Note that the performance of SCL methods ({\color{gray}gray}) is the same under different $N$ since no client data is used.
Cloudy is the server domain.
}\label{tab:bdd}
\vspace{-3mm}
\resizebox{0.85\textwidth}{!}{
\begin{tabular}{clcccccccccc}
\toprule
& \multirow{2.5}{*}{\bf Method} & \multicolumn{5}{c}{$N=3$} & \multicolumn{5}{c}{$N=9$} \\
\cmidrule(r){3-7} \cmidrule(l){8-12}
& & Cloudy & Overcast & Rainy & Snowy & \textbf{Total} & Cloudy & Overcast & Rainy & Snowy& \textbf{Total}  \\
\cmidrule(r){1-2} \cmidrule(r){3-6} \cmidrule(lr){7-7}\cmidrule(lr){8-11} \cmidrule{12-12}
\multirow{1}{*}{{Supervised Centralized}} & Backbone Tuned 
& $\color{gray}0.418$ & $\color{gray}0.447$ & $\color{gray}0.391$ & $\color{gray}0.387$ & $\color{gray}0.418$& $\color{gray}0.418$ & $\color{gray}0.447$ & $\color{gray}0.391$ & $\color{gray}0.387$ & $\color{gray}0.418$ \\ 
{Learning (SCL) }                                &  Backbone Frozen& $\color{gray}0.416$ & $\color{gray}0.449$ & $\color{gray}0.428$ & $\color{gray}0.429$ & $\color{gray}0.434$& $\color{gray}0.416$ & $\color{gray}0.449$ & $\color{gray}0.428$ & $\color{gray}0.429$ & $\color{gray}0.434$ \\
\cmidrule(r){1-2} \cmidrule(r){3-6} \cmidrule(lr){7-7}\cmidrule(lr){8-11} \cmidrule{12-12} 
\multirow{4}{*}{\shortstack[c]{Semi-Supervised \\Federated Learning\\(SSFL)}}            & FedAvg~\cite{fedavg} & \res{0.416} &\res{0.451} &\res{0.417} &\res{0.423} &\res{0.432}& $0.436$ & $0.462$ & $0.445$& $0.437$& $0.448$ \\
                                 & FedProx~\cite{fedprox} &\res{0.420}&\res{0.439}&\res{0.420}&\res{0.424}&\res{0.428}& $0.430$ & $0.456$ & $0.440$& $0.450$& $0.446$ \\
                                 & FedSTO~\cite{kim2024navigating} &\res{0.407}&\res{0.413}&\res{0.416}&\res{0.402}&\res{0.410}& $0.429$ & $0.457$ & $0.445$& $0.439$& $0.445$\\
                                 & \textbf{Ours} &\resb{0.468}&\resb{0.492}&\resb{0.483}&\resb{0.493}&\resb{0.486} & $\mathbf{0.465}$ & $\mathbf{0.486}$ & $\mathbf{0.478}$& $\mathbf{0.462}$& $\mathbf{0.475}$ \\
\bottomrule
\end{tabular}
}
\vspace{-2mm}
\end{table*}

\begin{table*}[t]
\begin{minipage}{0.6\textwidth}
    \centering
    \caption{\textbf{Evaluation of mAP on SODA10M dataset.} The performance of SCL methods ({\color{gray}gray}) is the same under different $N$ since no client data is used. Clear is the server domain.}
    \label{tab:soda}
    \vspace{-2mm}
    \resizebox{\textwidth}{!}{
    \begin{tabular}{clcccccccc}
    \toprule
    & \multirow{2.5}{*}{\bf Method} & \multicolumn{4}{c}{$N=3$} & \multicolumn{4}{c}{$N=9$} \\
    \cmidrule(lr){3-6} \cmidrule(lr){7-10}
    & & Clear  & Overcast & Rainy & \textbf{Total} & Clear  & Overcast & Rainy & \textbf{Total}  \\
    \cmidrule(r){1-2} \cmidrule(r){3-5}\cmidrule(lr){6-6} \cmidrule(lr){7-9} \cmidrule{10-10}
    
    \multirow{2}{*}{SCL} & Backbone Tuned&  \resg{0.220}&\resg{0.242}& \resb{\color{gray}0.378}&\resg{0.250}& \resg{0.220}&\resg{0.242}& \resg{0.378}&\resg{0.250}\\
                                     &Backbone Frozen& \resg{0.203}&\resg{0.245}& \resg{0.333}&\resg{0.238}&\resg{0.203}&\resg{0.245}& \resg{0.333}&\resg{0.238} \\
    \cmidrule(r){1-2} \cmidrule(r){3-5}\cmidrule(lr){6-6} \cmidrule(lr){7-9} \cmidrule{10-10}
    \multirow{4}{*}{SSFL}            & FedAvg~\cite{fedavg} & \res{0.223}&\res{0.253}& \res{0.359}&\res{0.253}& \res{0.222}&\res{0.255}& \res{0.366}&\res{0.255} \\
                                     & FedProx~\cite{fedprox} & \res{0.223}&\res{0.255}& \res{0.367}&\res{0.256}& \resb{0.223}&\res{0.253}& \res{0.362}&\res{0.254} \\
                                     & FedSTO~\cite{kim2024navigating} &\res{0.223}&\res{0.243}& \res{0.360}&\res{0.250}& \res{0.220}&\res{0.250}& \res{0.349}&\res{0.250} \\
                                     & \textbf{Ours} &\resb{0.235}&\resb{0.267}& \res{0.375}&\resb{0.267}& \resb{0.223}&\resb{0.263}& \resb{0.382}&\resb{0.261} \\
    \bottomrule
    \end{tabular}
    }
\end{minipage}
\hfill
\begin{minipage}{0.38\textwidth}
    \centering
    \caption{\textbf{Evaluation of mAP@50 on Cityscapes dataset.} The test data are from different \textit{cities} from the training data.}
    \label{tab:city}
    \vspace{-2mm}
    \resizebox{0.84\textwidth}{!}{
    \begin{tabular}{clcc}
    \toprule
    & {\bf Method} & $N=3$ & $N=9$ \\
    \cmidrule(r){1-2} \cmidrule(r){3-3} \cmidrule(lr){4-4}
    
    \multirow{2}{*}{SCL} & Backbone Tuned & {\resg{0.422} }& {\resg{0.422} } \\
                         & Backbone Frozen & {\resg{0.415}}&{\resg{0.415}}  \\
    \cmidrule(r){1-2} \cmidrule(r){3-3} \cmidrule(lr){4-4}
    \multirow{4}{*}{SSFL}  & FedAvg~\cite{fedavg}  & \res{0.411} & \res{0.423} \\
                            & FedProx~\cite{fedprox} & \res{0.417} & \res{0.441} \\
                            & FedSTO~\cite{kim2024navigating}  & \res{0.399} & \res{0.421} \\
                            & \textbf{Ours} & \resb{0.464} & \resb{0.455} \\
    \bottomrule
    \end{tabular}
    }
\end{minipage}
\vspace{-4mm}
\end{table*}

\subsection{Experimental Details}
\noindent\textbf{Comparison Methods.} Due to the constraints of our PSSFL setting, previous methods designed specifically for classification or requiring full-model training are \textit{not applicable} (\eg, FedBN~\cite{li2021fedbn}, FedPer~\cite{arivazhagan2019federated}, SemiFL~\cite{diao2022semifl}, SemiFed~\cite{lin2021semifed}). Therefore,
to demonstrate the superiority of our proposed method, we select several \textit{representative} methods with necessary adaptation for comparison. 
(a) To demonstrate the baseline performance with the server data only, we compare with two variants of \textit{supervised centralized learning (SCL)}, where we set the backbone tunable or frozen. 
(b) Under \textit{semi-supervised federated learning (SSFL)}, we adapt two widely used FL algorithms FedAvg~\cite{fedavg} and FedProx~\cite{fedprox} by combining them with the same semi-supervised learning. In addition, we compare with FedSTO~\cite{kim2024navigating}, which is a recent work tailored for semi-supervised federated object detection. Due to the frozen backbone in our setting, the selective training in FedSTO's original design is skipped.

\noindent\textbf{Dataset.} We simulate our proposed PSSFL setting in practical self-driving car scenarios for evaluation, where the clients, \ie, the vehicles, usually hold unlabeled images and constrained computational resources. 
We utilize the BDD100K~\cite{yu2020bdd100k}, SODA10M~\cite{han2021soda10m}, and Cityscapes~\cite{Cordts2016Cityscapes} datasets, which consist of image frames from driving video recordings under different conditions. Based on the available data annotation, we split BDD100K and SODA10M by the \textbf{weather} condition and Cityscapes based on the collected \textbf{city}.
In BDD100K, the detection categories are Pedestrian, Car, Bus, Truck, and Traffic signs. Similar to previous work~\cite{kim2024navigating}, we take $2,000$, $5,000$, $5,000$, and $8,000$ images under four weather conditions, \textit{Cloudy, Overcast, Rainy,} and \textit{Snowy}, and distribute $2,000$ Cloudy weather condition as labeled samples to the server and remaining images as unlabeled samples to clients.
Similarly, in SODA10M, we keep $2,000$ \textit{ Clear} weather conditions as labeled samples in the server and $10,000$, $8,000$ images under \textit{Overcast, Rainy} weather conditions to clients, where the object categories are Pedestrian, Cyclist, Car, Truck, Tram, and Tricycle. 
For Cityscapes, the categories are Person, Rider, Car, Truck, Bus, Train, Motorcycle, and Bicycle. Since images are collected from different cities without weather information, we split it in a \textit{domain-generalization style}, where the server, client, and test data are collected from different sets of cities. We select $2,000$ images from 18 cities as labeled samples 
for the server, and the remaining $18,000$ images from the other 23 cities as unlabeled samples to be uniformly distributed to clients. The $500$ test samples are from 3 different cities. We leave details for the dataset split in the \textit{{supplementary}}.

\noindent\textbf{Federated Learning Setting.} We conduct experiments in our proposed PSSFL setting as in Sec.~\ref{sec:problem} for all methods, where the server holds labeled data and the clients hold unlabeled data. We simulate the most challenging scenario, where server data and client data have \textit{no overlapping} in domains, and each client only owns dataset from one domain.
The image resolutions for servers and clients are set to $1280\times720$ and $640\times360$, respectively. We split each domain into several clients with $3$ and $9$ total clients. In each round, we will sample $33\%$ clients as online clients for local training. We further conduct experiments by scaling up the number of online and total number of clients in Sec.~\ref{sec:scale}.
Both the client and the server training epochs are $1$ in each round. We set the warm-up training epoch as $50$ and run $50$ FL rounds.

\noindent\textbf{Semi-Supervised Learning Setting.} We adopt a widely-used semi-supervised algorithm, Soft Teacher~\cite{xu2021end}, as the default training strategy for unsupervised learning for \textit{all methods}. More details are provided in the \textit{{supplementary}}.

\noindent\textbf{Model Architecture.} We use ViT-Adapter~\cite{chen2023vision} as the FM backbone, which consists of ViT pre-trained from DINOv2~\cite{oquab2023dinov2} and an adapter structure to support multi-scale feature extraction. The adapter structure is pre-trained on MS-COCO~\cite{lin2014microsoft}. The parameters in the FM backbone will remain frozen during the training. The trainable task head consists of an FPN neck and a Faster-RCNN head~\cite{ren2016faster} in our experiments. In FedMox, we apply sparse spatial MoE on both the RPN and ROI heads with the same number of experts $K$. On BDD100K, SODA10M, and Cityscapes, we set $K$ to 4, 3, and 3, respectively.


\subsection{Evaluation Results}
In order to comprehensively evaluate the methods, we conduct experiments with different focuses on datasets: (a)~The default setting evaluated with mean Average Precision at an IoU threshold of 50 (mAP@50) in BDD100K, (b)~The more comprehensive setting in SODA10M evaluated with mean Average Precision averaged among IoU threshold \textit{from 50 to 95} (mAP), and (c)~The domain generalization setting in Cityscapes evaluated on images collected in new cities. 

\noindent \textbf{Evaluation on BDD100K.} As shown in Table~\ref{tab:bdd}, on the BDD100K dataset, training only on server data (SCL) with Backbone Tuned yields better performance than with a Backbone Frozen on the server domain. However, due to overfitting, the Backbone Tuned shows a performance drop on unseen domains.
When client-side unlabeled data is incorporated (SSFL), FedProx and FedSTO fail to outperform the FedAvg baseline due to the limitations of the frozen backbone. This result highlights FedAvg’s strong capability under a pretraining setting, aligning with previous studies~\cite{fedopt,flpretrain}. Notably, our method, FedMox, outperforms all the other approaches across different client settings, benefiting from the enhanced adaptability of the Mixture of Experts (MoE) and the Soft Mixture strategy, which better balances updates between the server and clients.

\noindent \textbf{Evaluation on SODA10M.} The results on SODA10M show similar trends with slight differences, as shown in Table~\ref{tab:soda}. In SCL, the Backbone Tuned performs better than the Backbone Frozen, indicating that this is a more challenging scenario for models trained only on the task head. In this setting, our method consistently outperforms other approaches in most cases, further demonstrating the effectiveness of our approach.

\noindent \textbf{Evaluation on Cityscapes.} As shown in Table~\ref{tab:city}, FedMox still outperforms all comparison methods under the \textit{domain generalization} scenario, where test data is collected from unseen cities, indicating better generalizability of FedMox.

\begin{table}[t]
\centering
\caption{{
Scaling-up analysis on BDD100K and SODA10M (\eg, higher sampling rate $r$ or total number of clients $N$).}
}\label{tab:fl}
\vspace{-3mm}
\setlength{\tabcolsep}{3pt}
\resizebox{\linewidth}{!}{
\begin{tabular}{lccccccccc}
\toprule
\multirow{2.5}{*}{\bf Method} & \multicolumn{5}{c}{BDD100K, $r=67\%$} & \multicolumn{4}{c}{SODA10M, $N=100$}\\
\cmidrule(r){2-6} \cmidrule(l){7-10} 
 & Cloudy & Overcast & Rainy & Snowy & \textbf{Total} & Clear & Overcast & Rainy & \textbf{Total}\\
\cmidrule(r){1-1} \cmidrule(r){2-5} \cmidrule(lr){6-6}\cmidrule(lr){7-9} \cmidrule{10-10}
FedAvg &\res{0.446}&\res{0.463}&\res{0.443}&\res{0.443}&\res{0.452}&\res{0.218}&\res{0.252}&\res{0.354}&\res{0.251}\\
                                  FedProx &\res{0.428}&\res{0.457}&\res{0.435}&\res{0.440}&\res{0.443}&\resb{0.220}&\res{0.254}&\res{0.369}&\res{0.254}\\
FedSTO&\res{0.437}&\res{0.460}&\res{0.441}&\res{0.446}&\res{0.449}&\res{0.219}&\res{0.247}&\res{0.349}&\res{0.248}\\
\textbf{Ours} &\resb{0.456}&\resb{0.491}&\resb{0.484}&\resb{0.468}&\resb{0.478}&\resb{0.220}&\resb{0.261}&\resb{0.374}&\resb{0.258}\\
\bottomrule
\end{tabular}
}
\vspace{-5mm}
\end{table}

\subsection{Scaling-Up Analysis}\label{sec:scale}  
We evaluate our method in two expanded FL scenarios: (1) increasing the client sampling rate $r$ from 33\% to 67\% on BDD100K with \( N = 9 \), and (2) scaling up to \( N = 100 \) on SODA10M without reducing per-client sample counts.  

\noindent\textbf{Higher Client Sampling Rate.} On BDD100K with a 67\% sampling rate, our method achieves the highest total mAP@50 of 0.478, outperforming FedAvg (0.452), FedProx (0.443), and FedSTO (0.449). It also maintains strong performance across all weather conditions, benefiting from increased client participation.  

\noindent\textbf{More Clients.} On SODA10M with \( N = 100 \), our method attains a total mAP of 0.258, surpassing FedAvg (0.251), FedProx (0.254), and FedSTO (0.248). It excels in challenging conditions like Overcast (0.261) and Rainy (0.374), demonstrating scalability and robustness.  


\begin{table}[t]
\centering
\caption{\textbf{Ablation study on BDD100K.} HR indicates if high-resolution images are used for training on the server. FL indicates if the unlabeled data is leveraged via federated learning. MoE indicates if a MoE architecture is used. SM means the soft mixture.}\label{tab:ablation}
\vspace{-2mm}
\setlength{\tabcolsep}{3.5pt}
\resizebox{0.85\linewidth}{!}{
\begin{tabular}{cccccccccc}
\toprule
&\multirow{2.5}{*}{\bf HR} & \multirow{2.5}{*}{\bf FL} & \multirow{2.5}{*}{\bf  MoE} &\multirow{2.5}{*}{\bf  SM}  & \multicolumn{5}{c}{\bf mAP@50}\\
\cmidrule(l){6-10}
&& & &&Cloudy  & Overcast & Rainy & Snowy & \textbf{Total} \\
\midrule
\textcircled{\raisebox{-0.9pt}{1}}&
\cmark &\xmark & \xmark & \xmark& $0.416$ & $0.449$ & $0.428$ & $0.429$ & $0.434$\\
\textcircled{\raisebox{-0.9pt}{2}}&\cmark &\cmark & \xmark & \xmark& $0.436$ & $0.462$ & $0.445$& $0.437$& $0.448$ \\
\textcircled{\raisebox{-0.9pt}{3}}&\xmark &\cmark & \xmark & \xmark& \res{0.392} &\res{0.414} & \res{0.382} & \res{0.394} &\res{0.399}\\
\textcircled{\raisebox{-0.9pt}{4}}&\cmark &\xmark & \cmark & \xmark&  \res{0.419} &\res{0.448} & \res{0.398} & \res{0.419}& \res{0.426}\\
\textcircled{\raisebox{-0.9pt}{5}}&\cmark &\cmark & \cmark & \xmark&  \res{0.463} &\res{0.481} & \res{0.461} &\res{0.450} &\res{0.467}\\
\textcircled{\raisebox{-0.9pt}{6}}&\cmark &\cmark & \cmark & \cmark& $\mathbf{0.465}$ & $\mathbf{0.486}$ & $\mathbf{0.478}$& $\mathbf{0.462}$& $\mathbf{0.475}$\\
\bottomrule
\end{tabular}
}
\vspace{-5mm}
\end{table}

\section{Discussion}

To further demonstrate the effectiveness of our proposed method and gain more insights, we conduct more experiments on BDD100K with $9$ clients.

\subsection{Ablation Study}
We perform an ablation study on the BDD100K dataset to evaluate the contribution of three core components in our framework: (a) High-Resolution (HR) training data, (b) Federated Learning (FL) to leverage unlabeled client data, (c) the Mixture of Experts (MoE) model structure, and (d) the proposed Soft-Mixture (SM) strategy. Table~\ref{tab:ablation} presents the mean Average Precision (mAP) at an IoU threshold of 50 across different weather conditions: Cloudy, Overcast, Rainy, and Snowy, as well as the entire test set (Total).

\noindent \textbf{Effect of Federated Learning (FL).} As shown in \ic{1} and \ic{2} in Table~\ref{tab:ablation}, enabling FL allows clients to utilize their unlabeled data, which provides a slight improvement over fully supervised, centralized learning. This result highlights the potential of federated learning to enhance foundation model adaptation in our semi-supervised setting.

\noindent \textbf{Effect of High-resolution Images (HR).} A straightforward solution to address the resolution mismatch in PSSFL is to train with low-resolution images on the server to make the resolution match. To maintain the comparison fairness, we use \textit{high-resolution} images for evaluation. As shown in \ic{3}, there is a significant performance drop compared to \ic{2}, indicating the necessity of high-resolution images.

\noindent \textbf{Effect of Mixture of Experts (MoE).} Introducing the MoE structure improves performance on the server’s domain in a centralized setting by enhancing the task head, although it can lead to overfitting on client domains, as shown in \ic{1} and \ic{4}. In our semi-supervised federated setting shown in \ic{5}, however, MoE better addresses the resolution mismatch between server and clients and reduces overfitting by incorporating diverse, unlabeled client data.

\noindent \textbf{Effect of Soft Mixture (SM).} As shown in \ic{6}, applying the Soft Mixture achieves a better trade-off between the aggregated model and the server model, further boosting the performance compared to \ic{5}.


\begin{figure}[t] 
\centering 
\includegraphics[width=0.85\linewidth]{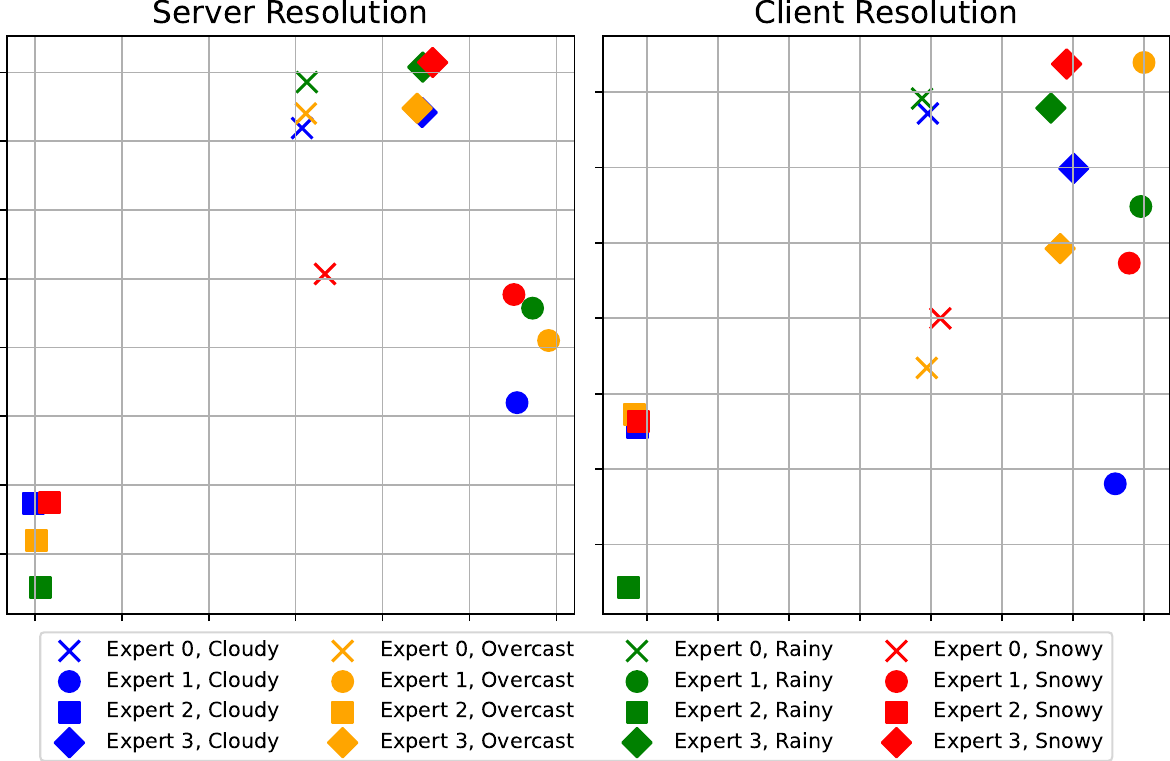} 
\caption{\textbf{Visualization of the expert routing for images with different resolutions.} Each scatter means the average location of the pixel routed to that expert. Consistency is shown between the server and client resolutions.} 
\label{fig:expert} 
\vspace{-5mm}
\end{figure}

\subsection{Analysis on MoE}\label{sec:moe_dis}
\textbf{Visualization of Experts.} To understand the knowledge captured by each expert, we visualize the routing maps for test samples and compute the \textit{average locations} assigned to each expert. Figure~\ref{fig:expert} shows the visualizations for images at both server and client resolutions. The results indicate that the spatial router clusters locations by expert rather than by domain, meaning pixels are routed to experts based on spatial position rather than origin domain. Furthermore, the expert distributions are similar across different resolutions, demonstrating that our spatial router effectively bridges the mismatched resolution between server and client data.


\noindent\textbf{Impact of Number of Experts.} As shown in Figure~\ref{fig:hyper} (left), increasing the number of experts does not lead to continuous performance improvement; instead, performance saturates a certain number of experts. This saturation occurs because additional parameters increase training difficulty, especially when labeled data is limited. Thus, in practical applications, selecting an optimal number of experts is crucial rather than assuming that a higher number will always yield better results. Over-parameterizing the model with too many experts can lead to diminishing returns and may even degrade performance due to training instability.

\subsection{Analysis on Soft Mixture}

\begin{figure}[t] 
\centering 
\includegraphics[width=0.99\linewidth]{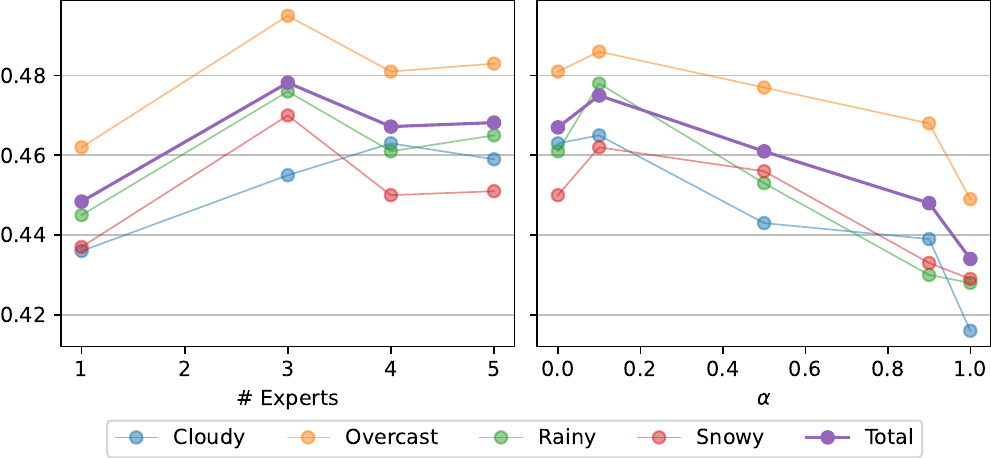} 
\caption{ Impact of soft mixture weights ($\alpha$) and the number of experts on mAP@50.} 
\label{fig:hyper} 
\vspace{-5mm}
\end{figure}

We evaluated the impact of the server and aggregated models in our soft mixture strategy by varying \(\alpha\), as shown in Figure~\ref{fig:hyper} (right). When \(\alpha = 0\), relying solely on the unsupervised aggregated model, sequential updates degrade precision. Conversely, \(\alpha = 1\), which depends only on supervised server training, results in poor generalization.  

Performance improves with increasing \(\alpha\), peaking before a slight decline. Notably, the model performs better when the aggregated model dominates (\(\alpha = 0.1\)) than when the server model does (\(\alpha = 0.9\)), suggesting that leveraging more information from the aggregated model can better balance precision and generalization in SSFL.

\subsection{Analysis on Communication and Computation}\label{sec:compute}

\begin{figure}[t]

        \centering
        \includegraphics[width=0.98\linewidth]{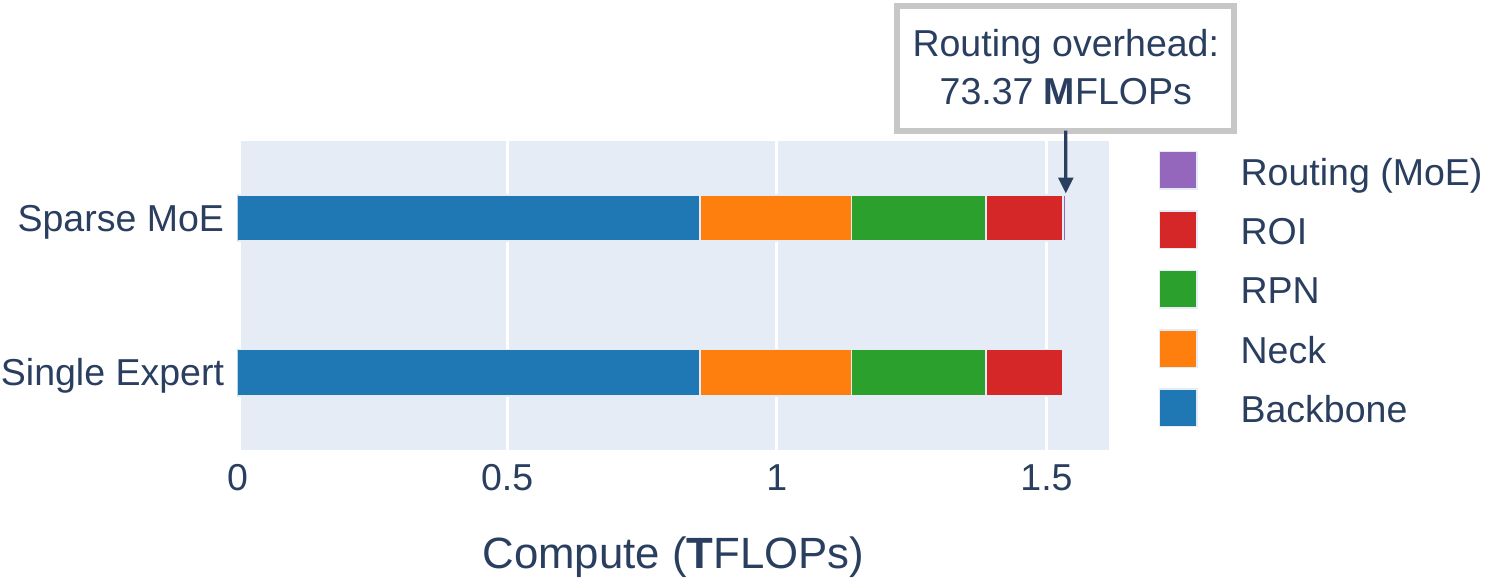}
\vspace{-3mm}
\caption{Computation Overhead of Sparse MoE and a single Expert. The additional Routing FLOPs is only 73.37M, which is negligible compared to other components.} 
\label{fig:compute} 
\vspace{-5mm}
\end{figure}

We demonstrate the communication and computation overhead for a Sparse MoE model with $K=3$.

\noindent\textbf{Communication.} Introducing additional task heads increases the communication overhead per client per round from 14.52M to 46.35M parameters. However, this increase is offset by the absence of backbone transmission.  
For reference, transmitting \textit{an entire model} requires 45.16M parameters, making the overhead \textit{comparable}.  

\noindent\textbf{Computation.} Compared to a single task head, the primary additional computation comes from the Routing operation, owing to our sparse MoE design. As shown in Figure~\ref{fig:compute}, the FLOPs required for Routing amounts to only 73.37M, which is negligible relative to other components.  



\section{Conclusion}
We introduced FedMox, a novel framework designed for our proposed practical semi-supervised federated learning setting, which reflects real-world constraints where clients have low-resolution, unlabeled data while the server uses high-resolution, labeled data to adapt the task head of a foundation model. 
FedMox addresses the core challenges of PSSFL: limited training flexibility from a frozen backbone, resolution mismatch between server and clients, and sequential updates between supervised and unsupervised learning.
Experiments on BDD100K, SODA10M, and Cityscapes show that FedMox outperforms existing SSFL methods, highlighting the effectiveness of FedMox in adapting foundation models to federated, semi-supervised environments, paving the way for resource-efficient and privacy-preserving model adaptation on edge devices. Beyond object detection, our proposed method is also adaptable to a broad range of computer vision tasks, which we plan to explore in future research.



{
    \small
    \bibliographystyle{ieeenat_fullname}
    \bibliography{main}
}
\clearpage
\setcounter{page}{1}
\maketitlesupplementary
\appendix

\section{Overview}
The supplementary material is structured as follows:
\begin{itemize}
    \item In Section~\ref{sec:alg}, we provide an algorithm for the overall setup of our proposed Practical Semi-Supervised Federated Learning (PSSFL).
    \item In Section~\ref{sec:am}, we discuss the applicability of related Federated Learning (FL) methods under our proposed PSSFL.
    \item In Section~\ref{sec:theory}, we present a theoretical analysis of soft mixture about the existence of an optimal $\alpha^*$.
    \item In Section~\ref{sec:exp}, we provide additional experiment details about the dataset, model, and training.
    \item In Section~\ref{sec:more_exp}, we analyze more possible designs for our Mixture-of-Expert (MoE) structure and discuss the privacy guarantee of PSSFL compared to traditional FL.
    \item In Section~\ref{sec:com}, we discuss the limitations and broader impact of our proposed framework.
\end{itemize}

\section{Algorithm of Our Proposed PSSFL setting} \label{sec:alg}

As shown in Algorithm~\ref{alg:semi_supervised_FL}, we provide a detailed algorithm for the entire process of PSSFL.

\section{Analysis on Applicable Methods for PSSFL} \label{sec:am}

We selected baselines based on their \textit{applicability} to PSSFL. As discussed in related work (\S 2) and comparison methods (\S 4.1), most existing FL approaches are unsuitable for PSSFL due to the following reasons:  

\noindent\textbf{Frozen backbone.} Methods requiring specific backbone designs (\eg, FedBN~\cite{li2021fedbn}) degrade to FedAVG~\cite{fedavg}.  

\noindent\textbf{Complex object detection tasks.} Methods designed for classification (\eg, FedProto~\cite{tan_fedproto_2022}, SemiFed~\cite{lin2021semifed}) cannot be directly applied.  

Additionally, approaches focused on efficient FM adaptation (\eg, LoRA~\cite{hu_lora_2021}) and federated optimization(\eg, MOON ~\cite{li2021model}) are orthogonal to our study and can be seamlessly integrated into our framework, making direct comparisons unnecessary.   


\section{Theoretical Analysis on Soft Mixture}\label{sec:theory}
This section provides a more theoretical analysis of our proposed soft-mixture strategy.

Consider an empirical risk minimization (ERM) problem for model weights $w$:
\begin{equation}
    \min_{w} L(w, \mathcal{D}_t),
\end{equation}
where $\mathcal{L}$ is the testing loss function and $\mathcal{D}_t$ is the test set.

\noindent \textbf{Assumptions.} We require the following assumptions:

\begin{assumption}
\textbf{(Loss Function Properties)} The Loss function for supervised training ($\mathcal{L}_s$), for unsupervised training ($\mathcal{L}_u$), and for testing ($\mathcal{L}$) are convex with respect to $w$ and differentiable.
\end{assumption}

\begin{assumption}
    \textbf{(Bounded Pseudo-labeling Error)} The pseudo-labeling error $\epsilon_p$ of foundation model adaption for unsupervised training is bounded by $\sigma$. That is, there exists $\sigma > 0$, $\mathbb{E}_{w}(||\epsilon_p||^2) < ||\sigma||^2$.
\end{assumption}

\begin{assumption}
    \textbf{(Data Diversity)} Unlabeled dataset $\mathcal{D}_u$ is more diverse than the labeled dataset $\mathcal{D}_s$, thus, has a smaller distribution discrepancy to the test dataset $\mathcal{D_t}$. That is, 
    \begin{equation}
        d_{\mathcal{H}}(\mathcal{D}_u, \mathcal{D}_t) < d_{\mathcal{H}}(\mathcal{D}_s, \mathcal{D}_t)
    \end{equation}
\end{assumption}

\begin{algorithm}[t]
\caption{\small Practical Semi-supervised Federated Learning}
\label{alg:semi_supervised_FL}
\vspace{1mm}
\begin{minipage}{0.9\linewidth}

\begin{algorithmic}
\small
\Require Server data $\mathcal{D}^s$, client data $\{\mathcal{D}_i^u\}_{i=1}^{N}$, pre-trained foundation model (FM), total rounds $T$, warm-up epochs $T_w$, number of online clients per round $M$.
\Ensure Trained global model $w^{(T)}$.

\State \textbf{Server Initialization:}
\State Freeze FM backbone and initialize detection head $w$.
\State Warm-up training on $\mathcal{D}^s$ for $T_w$ epochs to get $w^{(0)}$.

\For{$t = 0, \dots, T-1$} \Comment{Federated Learning Process}
    \State Server selects $M$ clients randomly.
    \For{each selected client $i \in \mathcal{M}$ \textbf{in parallel}}
        \State Receive global model $w^{(t)}$.
        \State Perform unsupervised training on $\mathcal{D}_i^u$.
        \State Send updated model $w_i^{(t+1)}$ back to server.
    \EndFor
    \State Server aggregates $\{w_i^{(t+1)}\}$.
    \State Perform high-resolution training on $\mathcal{D}^s$ to get $w^{(t+1)}$.
\EndFor

\State \textbf{Evaluation:} Test final model $w^{(T)}$ on a hold-out test set.

\vspace{1mm}
\end{algorithmic}
\end{minipage}
\end{algorithm}

\begin{remark} The model with the soft mixture $w_\alpha$ has a combined risk respect to $\alpha \in [0,1]$:
\begin{multline}
    \mathcal{L}(w_\alpha; \mathcal{D}_t) \le \alpha \left[ \hat{\mathcal{L}}(w) + d_{\mathcal{H}}(\mathcal{D}_s,\mathcal{D}_t) \right ] \\ + (1-\alpha) \left[ \hat{\mathcal{L}}(\bar w) + d_{\mathcal{H}}(\mathcal{D}_u,\mathcal{D}_t)\right]\\ + \mathcal{O}\left( \frac{1}{\sqrt{n_s+n}} \right) 
\end{multline}
\end{remark}

\begin{theorem} There exist an optimal $\alpha^*$ to achieve a better trade-off, yielding a more optimal $ \mathcal{L}(w_\alpha; \mathcal{D}_t)$ respect to 
\begin{equation}
    \alpha^* = \arg \min_{\alpha\in[0,1]}\mathcal{L}(w_\alpha; \mathcal{D}_t),
\end{equation}
where $\frac{\partial\mathcal{L}(w_{\alpha^*}; \mathcal{D}_t)}{\partial{\alpha^*}} = 0$.
    
\end{theorem}

\begin{figure}[t]
    \centering
    \includegraphics[width=\linewidth]{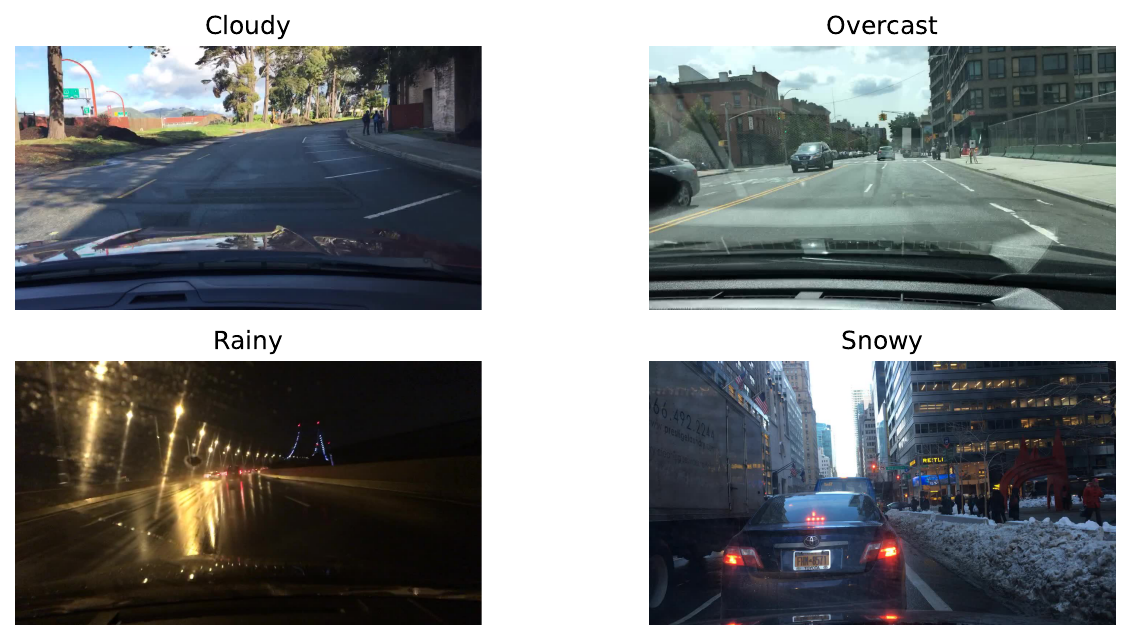}
    \caption{Visualization for images in BDD100K in each domain}
    \label{fig:bdd}
\end{figure}

\begin{figure}[t]
    \centering
    \includegraphics[width=\linewidth]{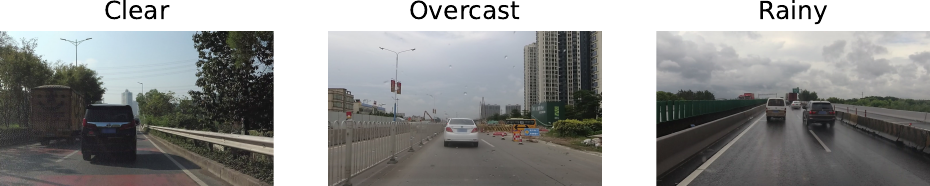}
    \caption{Visualization for images in SODA10M in each domain}
    \label{fig:soda}
\end{figure}

\begin{figure}[t]
    \centering
    \includegraphics[width=\linewidth]{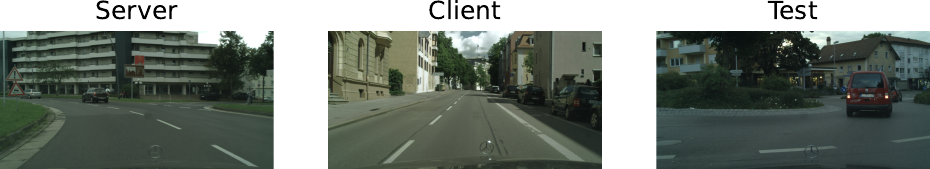}
    \caption{Visualization for images in Cityscapes in server, client, and test dataset}
    \label{fig:city}
\end{figure}

\section{Additional Experimental Details}\label{sec:exp}
\subsection{Visualization of Samples in Each Dataset}
As shown in Figures~\ref{fig:bdd},~\ref{fig:soda}, and~\ref{fig:city}, we present examples of images in each dataset with related domain or training split information for our experiments.

\begin{table*}[t]
\centering
\caption{Evaluation on BDD100K dataset with different scales of objects.
}\label{tab:scale}
\vspace{-3mm}
\resizebox{\textwidth}{!}{
\begin{tabular}{lccccccccccccccc}
\toprule
\multirow{2.5}{*}{\bf Method} & \multicolumn{5}{c}{mAP$_\text{small}$} & \multicolumn{5}{c}{mAP$
_\text{medium}$} & \multicolumn{5}{c}{mAP$_\text{large}$}\\
\cmidrule(r){2-6} \cmidrule(l){7-11}  \cmidrule(l){12-16}
 & Cloudy & Overcast & Rainy & Snowy & \textbf{Total} & Cloudy & Overcast & Rainy & Snowy& \textbf{Total}& Cloudy & Overcast & Rainy & Snowy& \textbf{Total}  \\
\cmidrule(r){1-1} \cmidrule(r){2-5} \cmidrule(lr){6-6}\cmidrule(lr){7-10} \cmidrule{11-11} \cmidrule(lr){12-15} \cmidrule{16-16}
FedAvg~\cite{fedavg} &\res{0.081} &  \res{0.100}&  \res{0.062}&  \res{0.074}&  \res{0.083}&  \res{0.258}&  \res{0.264}&  \res{0.235}&  \res{0.217}&  \res{0.248}&  \res{0.466}&  \res{0.429}&  \res{0.391}&  \res{0.402}&  \res{0.423}\\
                                  FedProx~\cite{fedprox} &\res{0.084} &  \res{0.102}&  \res{0.058}&  \resb{0.083}&  \res{0.086}&  \res{0.261}&  \res{0.264}&  \res{0.232}&  \res{0.230}&  \res{0.250}&  \res{0.491}&  \res{0.435}&  \res{0.398}&  \res{0.424}&  \res{0.437}\\
FedSTO~\cite{kim2024navigating}&\res{0.087} &  \res{0.104}&  \res{0.065}&  \res{0.071}&  \res{0.086}&  \res{0.253}&  \res{0.262}&  \res{0.225}&  \res{0.231}&  \res{0.247}&  \res{0.458}&  \res{0.426}&  \res{0.375}&  \res{0.397}&  \res{0.416}\\
\textbf{Ours} & \resb{0.099} &  \resb{0.112}&  \resb{0.070}&  \res{0.076}&  \resb{0.094}&  \resb{0.276}&  \resb{0.285}&  \resb{0.249}&  \resb{0.250}&  \resb{0.269}&  \resb{0.501}&  \resb{0.474}&  \resb{0.413}&  \resb{0.437}&  \resb{0.460}\\
\bottomrule
\end{tabular}
}
\vspace{-2mm}
\end{table*}

\subsection{Implementation Details}
\subsubsection{Dataset} 
\noindent \textbf{BDD100K~\cite{yu2020bdd100k}.} There are $1,500$ test samples in the test set, where $300$, $600$, $300$, and $300$ samples are under Cloudy, Overcast, Rainy, and Snowy weather conditions, respectively. The test samples are randomly sampled from the original validation set in BDD100K. We assign $1$ and $3$ clients per domain for $N=3$ and $N=9$, respectively.

\noindent \textbf{SODA10M~\cite{han2021soda10m}.} There are $2,300$ test samples in the test set, where $1,000$, $1,000$, and $300$ samples are under Clear, Overcast, and Rainy weather conditions, respectively. The test samples are randomly sampled from the original validation set of labeled data in SODA10M. We assign $(2, 1)$ and $(6, 3)$ clients under (Overcast, Rainy) for $N=3$ and $N=9$, respectively.

\noindent \textbf{Cityscapes~\cite{Cordts2016Cityscapes}.} The original $500$ samples from the fine-grained labeled data in Cityscapes are directly used as the test set since there is no weather information provided.

\subsubsection{Model} 
Our model is implemented based on the Faster R-CNN~\cite{ren2016faster} detector in MMDetection~\cite{mmdetection}. The details of the backbone, neck, and detection head are provided as follows:

\noindent \textbf{Backbone.} Our ViT-Adapter-Small backbone~\cite{chen2023vision} is implemented based on their \href{https://github.com/czczup/ViT-Adapter}{original} implementation. Specifically, our backbone has an image size of $518$, patch size of $14$, and embedding dim of $384$ with $6$ heads.

\noindent \textbf{Neck.} We employ a Feature Pyramid Network neck with $5$ out channels with a dimension of $256$.

\noindent \textbf{Detection Head.} For the Region Proposal Network (RPN), we use the \texttt{MaxIoUAssigner} for anchor assignment with a positive IoU threshold of $0.7$ and a negative threshold of $0.3$. The RPN sampler selects $256$ samples per image, maintaining a 1:1 ratio of positive to negative samples. RPN proposals undergo Non-Maximum Suppression (NMS) with an IoU threshold of $0.7$, retaining up to $2,000$ proposals before NMS and $1,000$ proposals per image after NMS.
For the R-CNN stage of the ROI head, the \texttt{MaxIoUAssigner} is used with both positive and negative IoU thresholds set to $0.5$. The R-CNN sampler selects $512$ RoIs per image, with a positive fraction of $0.25$, and ground truth boxes are added as proposals during sampling. The mask size for training is set to $28$, and weights for positive samples are kept at the default of $-1$. 

\noindent \textbf{Mixture of Experts (MoE) Design of the detection head.} MoE is applied to the region proposal network (RPN) head and the bounding box (BBOX) head of the region of interest (ROI) head separately. Both MoEs have a sparse top-1 router. The MoE for the RPN head is a spatial router due to the different sizes of input feature maps, while the MoE for the ROI head is a traditional router with a fixed dimension of the ROI feature size. 

\subsubsection{Training}

\noindent \textbf{Semi-supervised Learning with Soft Teacher~\cite{lin2014microsoft}.} During training, teacher proposals are not used. Pseudo-labeling thresholds are set at $0.5$ for the initial score, $0.9$ for RPN pseudo labels, $0.9$ for classification pseudo labels, and $0.02$ for regression pseudo labels. Data augmentation includes jittering applied $10$ times with a scale of $0.06$, and no minimum size is enforced for pseudo boxes.
An unsupervised weight of $4.0$ balances the supervised and unsupervised losses. During testing, inference is performed using the student model.

\noindent \textbf{Federated Learning.} Our federated learning simulation is based on the COALA~\cite{zhuangcoala} platform. We use the AdamW~\cite{loshchilov2018decoupled} optimizer with a learning rate of $0.0001$ and a weight decay of $0.05$ to train the model. For learning rate scheduling, we adopt a cosine annealing strategy with a warmup phase lasting $5$ epochs. This setup ensures stable optimization and smooth convergence throughout both the client and server training.

\subsubsection{Hyperparameters for Comparison Methods}
\noindent \textbf{FedProx~\cite{fedprox}.} We set the hyperparameter $\mu=0.001$ for regularization.

\noindent \textbf{FedSTO~\cite{kim2024navigating}.} We set the hyperparameter $\mu=0.001$ for orthogonal enhancement regularization.

\section{Additional Discussion}\label{sec:more_exp}

\subsection{Impact of Different Design of MoE}
To validate the effectiveness of our MoE design, we tested different MoE configurations and training methods, as shown in Table~\ref{tab:moe}. An intuitive alternative to our spatial MoE design is manually assigning each expert to a specific domain and training the MoE on the server (Domain-Assigned). However, due to the resolution mismatch, the MoE trained on the server performs poorly when deployed on clients.
Another approach leverages the server's computational resources by training a dense mixture of all experts (Dense Experts) or a joint training of sparse and dense experts (Dense + Top-1) on the server side. However, our experiments show that these methods, attributed to the discrepancy between server and client training processes, perform worse than a consistent training strategy across both server and client, where only the top expert is trained sparsely (Top-1). 
\begin{table}[t]
\centering
\caption{Performance of different MoE designs.}\label{tab:moe}
\vspace{-2.5mm}
\resizebox{\linewidth}{!}{
\begin{tabular}{lccccc}
\toprule
\multirow{2.5}{*}{\bf  MoE Designs}  & \multicolumn{5}{c}{\bf mAP@50}\\
\cmidrule(lr){2-6}
 &Cloudy  & Overcast & Rainy & Snowy & Total \\
\midrule
No MoE& $0.436$ & $0.462$ & $0.445$& $0.437$& $0.448$ \\
\midrule
Domain-Assigned& \res{0.435}& \res{0.467}& \res{0.425}& \res{0.444}& \res{0.448}\\
Dense Experts& \res{0.444}& \res{0.479}& \res{0.445}& \resb{0.465}& \res{0.462}\\
Dense + Top-1& \res{0.456} & \res{0.470}& \res{0.440}& \res{0.458}& \res{0.459} \\
Top-1 (Ours)& \resb{0.463} &\resb{0.481} & \resb{0.461} &\res{0.450} &\resb{0.467}\\
\bottomrule
\end{tabular}
}
\vspace{-5.5mm}
\end{table}

\subsubsection{Implementation Details of MoE Designs}
\noindent \textbf{Domain-assigned.} We pre-assign each client with a corresponding expert based on the domain of its local data. In the aggregation stage, each expert is aggregated separately. That is, no inter-expert aggregation happens. The supervised training will be conducted on all experts and a traditional router. Note that the router is only trained on the server, since there is only one expert on each client.

\noindent \textbf{Dense + Top-1.} To jointly train the MoE with dense routing and top-1 routing, we switch the routing strategy between dense and sparse (top-1) for each training step.

\subsection{Privacy Discussion}
Our framework doesn't add additional privacy risk beyond traditional FL and is compatible with existing FL defense methods. Besides, existing attacks for FL are under strong assumptions where \textit{classification objective} and \textit{small batch size}, making it relative extremely challenging for complex tasks such as object detection. 

\section{Limitation and Potential Negative Societal Impacts}\label{sec:com}

\noindent \textbf{Limitations.} Communication cost is another significant challenge in FL. As discussed in previous MoE-based frameworks, the communication cost of MoE increases with the number of experts \(K\). However, in our setting, communication is not a dominant bottleneck compared to computational constraints, as only the task head is transmitted in each round. Although various methods exist to further compress communication costs~\cite{dusenberry2020efficient}, this aspect is outside the scope of this paper and will be explored in future work.

\noindent \textbf{Potential Negative Impacts.} While our framework aims to address challenges in semi-supervised federated learning (SSFL), it has potential societal risks. It could be misused in surveillance systems, exacerbate inequalities due to uneven access to resources, or amplify biases if the training data is unrepresentative. 
Despite its privacy-preserving design, vulnerabilities to adversarial attacks like data reconstruction could compromise user privacy. Additionally, training large models may contribute to environmental concerns due to high energy consumption. Careful consideration of ethical use, robust privacy mechanisms, and dataset fairness is essential to mitigate these risks.

\end{document}